# 基于双侧通道特征融合的三维姿态迁移网络

刘珏 [1,2)]

[1)] (东南大学自动化学院, 南京　210096)
[2)] (东南大学复杂工程测量与控制教育部重点实验室, 南京　210096)
(liu_jue@seu.edu.cn)

**摘　要:** 为了解决现有方法中, 姿态特征在前向传播过程中的姿态失真问题, 本文提出了一种基于双侧通道特征融合的姿态迁移网络. 首先, 通过姿态编码器从源网格中提取固定长度的姿态编码, 并将其和目标顶点组合成混合特征; 然后, 设计了一种特征融合自适应实例归一化模块, 该模块可同时处理姿态和身份特征, 使姿态特征在逐层前向传播中得到补偿, 从而解决姿态失真问题; 最后, 使用该模块构成的网格解码器, 逐步将姿态迁移到目标网格上. 在 SMPL、SMAL、FAUST 和 MultiGarment 数据集上的实验结果表明, 本文的方法在保持较小网络结构的同时, 具有更强的姿态迁移能力和更快的收敛速度, 成功解决了姿态失真问题, 同时能够适应不同顶点数的网格. 本文方法的代码可用: https://github.com/YikiDragon/DSFFNet

**关键词:** 姿态迁移; 形变迁移; 卷积神经网络; 条件归一化

**中图法分类号:** TP391.41　　　　**DOI:** 10.3724/SP.J.1089.202*.论文编号

## DSFFNet: Dual-Side Feature Fusion Network for 3D Pose Transfer

Jue Liu [1,2)]

[1)] (*School of Automation, Southeast University, Nanjing,* 210096)
[2)] (*Key Laboratory of Measurement and Control of Complex Systems of Engineering, Ministry of Education, Nanjing,* 210096)

**Abstract:** To solve the problem of pose distortion in the forward propagation of pose features in existing methods, this paper proposes a Dual-Side Feature Fusion Network for pose transfer (DSFFNet). Firstly, a fixed-length pose code is extracted from the source mesh by a pose encoder and combined with the target vertices to form a mixed feature; Then, a Feature Fusion Adaptive Instance Normalization module (FFAdaIN) is designed, which can process both pose and identity features simultaneously, so that the pose features can be compensated in layer-by-layer forward propagation, thus solving the pose distortion problem; Finally, using the mesh decoder composed of this module, the pose are gradually transferred to the target mesh. Experimental results on SMPL, SMAL, FAUST and MultiGarment datasets show that DSFFNet successfully solves the pose distortion problem while maintaining a smaller network structure with stronger pose transfer capability and faster convergence speed, and can adapt to meshes with different numbers of vertices. Code is available at https://github.com/YikiDragon/DSFFNet

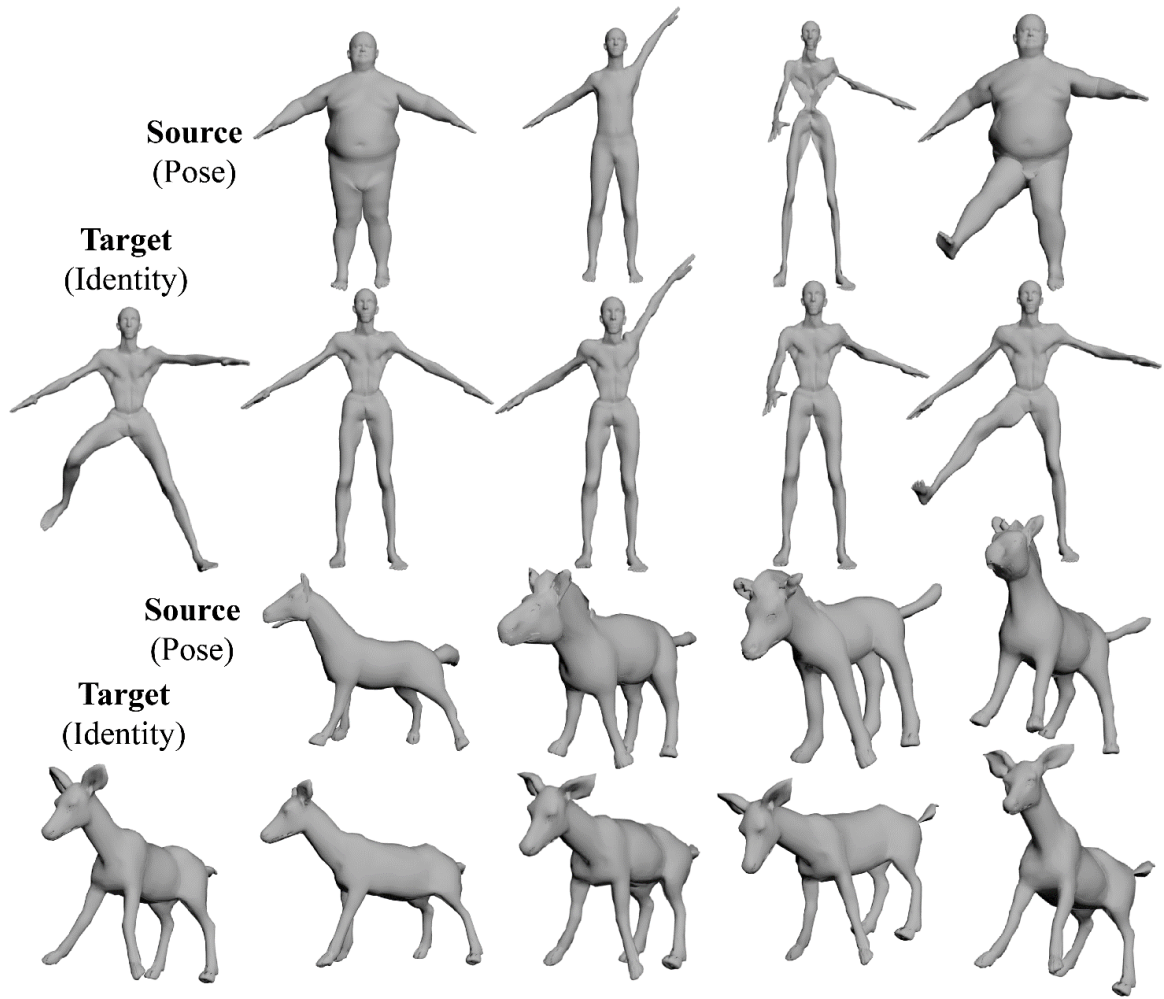

图 1 DSFFNet 的姿态迁移结果. "Source"和"Target"是形变迁移的习惯表示. 上面两行是来自 SMPL 的人体身份和姿态, 下面两行是来自 SMAL 的动物身份和姿态.

## 1 引 言

三维姿态迁移逐渐成为计算机图形学和视觉领域的研究热点. 它源自以往的形变迁移技术, 经过大量研究已证明其卓越性能. 过去的形变迁移方法通常需要大量额外的输入, 如顶点之间的对应关系[1]、三维模型的骨架及权重信息[2]、辅助网格[3]、关键点标注[4]等. 然而, 在实际应用中, 这些信息往往难以获取. 因此, Wang 等人首次提出了神经姿态迁移(NPT)的概念[5], 可以直接将源网格的姿态迁移到目标网格, 无需参考网格. 随后, Song 等人在 NPT 的基础上提出了 3D-CoreNet[6,7], 改进了条件归一化模块, 并使用最优传输方法解决了源和目标的对应问题, 从而进一步提高了姿态迁移精度. 尽管 NPT 和 3D-CoreNet 都为三维姿态迁移领域做出了重要贡献, 但它们在精度、模型大小和训练难度等方面仍存在一些不足之处.

NPT 作为一项开创性工作, 摒弃了形变迁移中的参考网格, 直接以源网格和目标网格作为输入. 它通过空间自适应实例归一化模块(SPAdaIN)学习网格姿态的空间变化, 从而实现对目标网格的形变. 然而, SPAdaIN 仅在前向传播过程中对姿态特征进行解码, 而侧通道仅学习目标网格的特征, 这导致 NPT 的网格解码器更侧重于保留目标网格的特征, 同时导致姿态特征在多次前向传播后出现失真问题.

NPT 的 origin 变体采用了二维矩阵作为姿态编码, 矩阵大小随源网格顶点数量变化, 这使得模型无法处理源网格和目标网格顶点数量不同的情况. 而 NPT 的 maxpool 变体以固定长度的一维向量作为姿态编码, 虽然可以适用于顶点数不同的源网格和目标网格, 但相比 origin 变体, 其精度有所下降.

3D-CoreNet 采用了 NPT(origin)的姿态编码器来提取源网格和目标网格的姿态编码矩阵, 然后通过解决两者之间的最优传输问题获得源和目标

的传输矩阵．接着，它使用传输矩阵和源网格来生成只包含姿态信息的扭曲网格，并通过弹性实例归一化模块(ElaIN)将目标网格的身份特征融合到扭曲网格中，最终得到重建网格．然而，3D-CoreNet 与 NPT 相比，仅将姿态特征的表现形式从编码变成扭曲网格，而在解码过程中仍无法避免姿态失真问题．

此外，计算传输矩阵的过程涉及大规模矩阵乘法，这不仅降低了前向传播速度，还导致了庞大的反向传播计算图，增加了模型训练的难度．而 ElaIN 在 SPAdaIN 的基础上引入了大量学习参数，导致模型占用大量内存．尽管它提高了迁移精度，但这并不能弥补其在模型大小和训练成本方面的不足．

避免前向传播过程中的姿态失真问题是提高姿态迁移精度的关键．无论是 NPT 还是 3D-CoreNet，它们的姿态特征仅在前向通道中传播，容易引发失真．解决这一问题的直接方法是通过在解码器的每个条件归一化模块中引入学习单元，使其能够学习未失真的姿态特征．这些学习单元可以构建一个侧通道，以便每个条件归一化模块都能够利用未失真的姿态特征来补偿前向传播过程中产生的失真，从而解决姿态失真问题．

本文提出了一种基于双侧特征融合的三维姿态迁移方法(DSFFNet)．图 1 展示了 DSFFNet 的一些示例．受到语义图像合成方法 SEAN[8]的启发，本文通过在 SPAdaIN 模块上引入了一个用于学习姿态特征的侧通道，设计了特征融合自适应实例归一化模块(FFAdaIN)．新增通道的输入是由姿态编码向量与目标顶点构成的混合特征，这确保了 DSFFNet 能够适应不同顶点数量的网格．混合特征侧通道与目标顶点侧通道构成了 FFAdaIN 的双侧通道结构．FFAdaIN 从两个侧通道学习姿态特征和身份特征，同时设置自适应系数，在训练过程中自动调整姿态与身份特征的比例，并将它们转化为前向通道中的反归一化参数，从而对前向传播过程中的姿态特征和身份特征进行补偿．这种结构同时避免了姿态和身份的特征失真问题，使模型的训练过程更快收敛，更好地拟合 Ground Truth. 此外，DSFFNet 在已见过的和未见过的姿态迁移精度方面也有显著提高．

本文的主要贡献总结如下：

1. 提出了 DSFFNet：一种用于三维姿态迁移的双侧通道特征融合网络．

2. 采用了固定长度的一维向量作为姿态编码，使 DSFFNet 能够灵活适用于不同顶点数量的网格．

3. 创新性地提出了特征融合自适应实例归一化模块 FFAdaIN，成功解决了以往方法在前向传播过程中产生的姿态特征失真问题．

4. DSFFNet 提高了姿态迁移精度的同时，显著加速了训练速度，同时保持了相对较小的网络体积．

## 2 相关工作

### 2.1 形变迁移

三维姿态迁移源自三维形变迁移领域．Sumner 等人首次提出了经典的形变迁移算法(DT)[1]．该方法主要包括对应关系生成和网格形变两个步骤．它使用稀疏矩阵表示网格之间的形变过程，通过最小二乘法将参考网格变换为与目标网格相近的形状．然后，通过 KDTree 和源面与目标面的法向量夹角来生成源网格与目标网格的对应关系，最后再次应用最小二乘法将形变迁移到目标网格上．然而，由于 DT 算法需要生成网格之间的密集对应关系，因此源网格与目标网格的顶点数量必须相同．

为了解决这个问题，Ben-Chen 等人首次提出了一种方法，不直接将形变应用到三维模型，而是将形变应用于包围模型的空间区域，即"笼"网格[4]．对于三维网格，笼由网格表面生成，而对于三维点云，笼由泊松表面重建法[9]构建．然后，笼的顶点经过简化和膨胀后，作为最终的"笼"来控制源网格或点云．然而，"笼"之间的形状差异容易导致错误的形变．

近年来，基于 PointNet 改进的点云处理方法已广泛应用于三维形变迁移领域[10,11]．Yifan 等人提出了一种基于深度学习的"笼"生成方法[12]，利用设计的编解码器结构将同一个球型网格分别转换为适用于源网格和目标网格的"笼"．然后，通过 MVC 函数[13]构建"笼"与其包裹的网格之间的映射关系，从而实现源网格的"笼"应用到目标网格上进行形变迁移．随后，Sung 等人提出使用隐含空间中的向量来编码形变[14]，并结合点云的形状特征进行解码以得到最终的点云．Liao 等人提出使用图卷积网络来预测网格的蒙皮权重[15]，然后使用 ICP

配准算法[16]计算网格各部分的变换矩阵,最后使用 LBS 算法[17]来获得最终的网格. 然而,形变迁移通常需要三个网格的输入:源网格、参考网格和目标网格,而在实际应用中,获取参考网格通常是困难的.

### 2.2 姿态迁移

三维网格姿态迁移是一项极具挑战性的三维生成任务. Wang 等人首次提出了神经姿态迁移(NPT)的概念[5]. 受到二维图像风格迁移方法的启发,他们引入了SPAdaIN模块,该模块可以将目标网格的身份特征迁移到源的姿态上,从而间接实现了姿态迁移并产生了惊人的效果. Song 等人提出了 3D-CoreNet[6,7],其中的 ElaIN 模块将前向传播特征与侧面输入特征加权求和,并将结果用作反归一化参数. 此外,他们采用最优传输方法来解决目标网格和源网格之间的密集对应问题. 然而,由于最优传输计算涉及大矩阵乘法,因此该方法占用大量内存并降低计算速度.

Chen 等人尝试将注意力机制引入姿态迁移网络[18]. 然而巨大模型的性能提升十分有限. 他们还提出使用生成对抗网络(GAN)来解决姿态迁移问题[19],这个方法借鉴了二维图像多特征解耦的思想[20]. 他们通过编码器将网格编码为内在特征码和外在特征码,其中内在特征码包含网格的身份特征,外在特征码包含网格的姿态特征. 然后,他们将目标网格的内在特征码与源网格的外在特征码相结合,送入生成器以输出的重建后的网格. 类似的思想也在其它工作中有所体现[21-23],这些方法将点云编码为内在特征、外在特征和自身旋转特征,通过控制外在特征来控制重建点云的姿态. 然而,这些方法通常因为复杂的模型结构和损失函数而难以训练.

### 2.3 条件归一化

条件归一化在任意风格迁移和语义图像合成中被广泛应用. 任意风格迁移的典型方法是AdaIN[24],它是基于 InstanceNorm[25]的改进版本. 通常,InstanceNorm 被视为一种图像风格的归一化方法,而 AdaIN 则利用风格图像各通道的均值和方差来对已归一化后的图像进行反归一化,从而实现风格迁移. 然而,AdaIN 不包含可学习的参数,因此其迁移性能有限. Chandran 等人提出了AdaConv 模块[26],它扩展了 AdaIN,使其能够提取风格图像中的局部特征,从而生成更加细致的图像. Liu 等人提出了在 AdaIN 基础上增加注意力机制的方法[27],虽然略微提高了模型的精度和泛化能力,但也需要较大内存才能进行训练.

在语义图像合成中,条件归一化通常在张量的 Batch 维度进行. SPADE[28]设计了三个卷积层,用于提取语义映射图的特征,然后使用这些特征对经过 BatchNorm 后的数据进行反归一化,以获得合成图像. 然而,在反归一化过程中,SPADE 仅使用了语义图,而忽视了源图像的风格特征. Zhu 等人提出的 SEAN[8]增加了对源图像风格特征的编码,并设计两个可学习参数,用于对风格特征和语义特征进行加权求和,从而使合成图像能够更好地保留源图像的风格. 本文方法受到了 AdaIN 和 SEAN 的启发.

## 3 方  法

本文提出的 DSFFNet 的总体架构如图 2 所示. 该模型主要由两部分组成:第一部分使用姿态编码器从源网格中提取姿态编码向量,然后将姿态编码向量与目标顶点进行 Concat 操作,构成混合特征. 第二部分由 FFAdaIN Resblock 构成的网格解码器,以目标顶点作为前向输入,混合特征和目标顶点作为侧面输入. 两个侧通道从混合特征和目标顶点中提取特征融合参数,并对前向通道的张量进行反归一化. 这一过程使得在逐层传播过程中,源网格的姿态被迁移到目标网格上. 接下来,本文将对姿态编码器和网格解码器进行详细介绍.

### 3.1 姿态编码器

姿态编码器的详细架构如图 3(a)所示. 本文沿用了 NPT[5]的设计,但在输出方面使用了 maxpool 层,以提取一维向量作为姿态编码. 姿态编码器包含了三组卷积单元,每个卷积单元按照 Conv1d-IN-ReLU 的方式排列,卷积核大小为 1×1. 以点集形状为 $N_{src} \times 3$ 的源网格为例,每个卷积单元输出张量的第一维度保持为 $N_{src}$,第二维度逐层递增. 在最后一个卷积单元,输出的张量大小为 $N_{src} \times 1024$,通过一个 maxpool 层提取出长度为 1024 的姿态编码向量 $z_{po}$. 根据目标顶点个数 $N_{tgt}$,姿态编码 $z_{po}$ 通过 repeat 操作转换为 $N_{tgt} \times 1024$ 大小,然后与目标顶点 $v_{id}$ 进行 Concat 操作得到混合特征 $f_{mix}$. 由于姿态编码 $z_{po}$ 的形状不会随源网格的顶点数量变化而改变,因此构建出的混合特征 $f_{mix}$ 大小可以固定为 $N_{tgt} \times 1027$,可直接用于后续解码过程. 相反,如果去除 maxpool 层,使用 $N_{src} \times 1024$

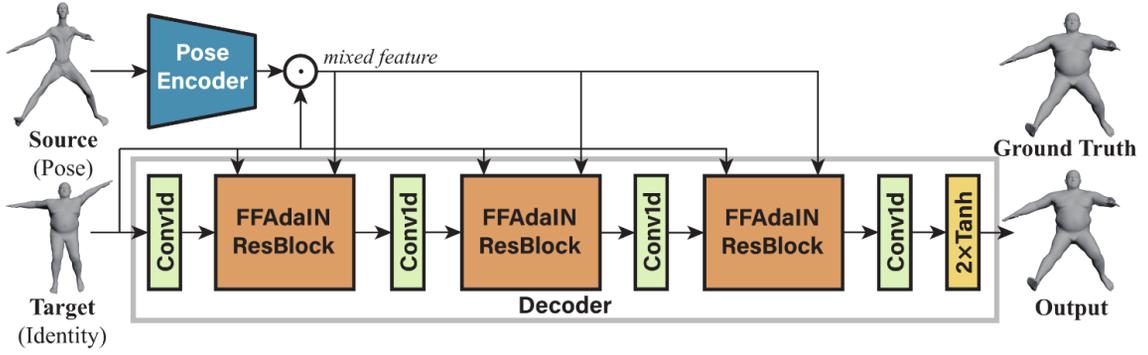

图 2 DSFFNet 的整体架构. 以源网格和目标网格作为输入, 姿态编码 $z_{po}$ 由姿态编码器从源网格提取. 混合特征 $f_{mix}$ 由目标顶点 $v_{id}$ 和 $z_{po}$ 构成. 网格解码器在 $f_{mix}$ 和 $v_{id}$ 指导下生成输出网格. 符号 ⊙ 表示 Concat 操作.

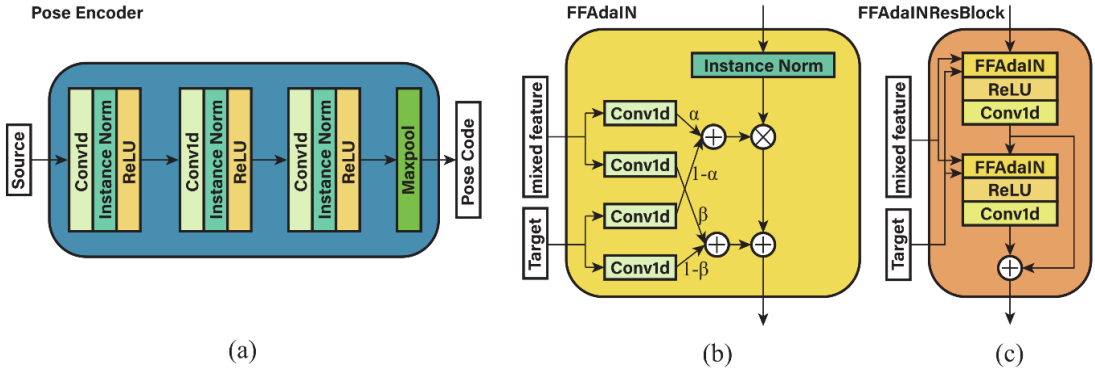

图 3 网络组件的具体架构. (a) 姿态编码器结构; (b) FFAdaIN 结构; (c) FFAdaIN Resblock 结构.

的二维矩阵作为姿态编码, 那么当 $N_{src} \neq N_{tgt}$ 时, 就无法将 $z_{po}$ 与 $v_{id}$ 通过 Concat 操作构建成混合特征. 因此, 采用带有 maxpool 结构的设计的最主要目的是使得 DSFFNet 能够适用于不同顶点数量的网格. 这种结构在 NPT 中已经被证明可以有效提取网格的姿态信息, 同时长度为 1024 的一维姿态编码向量足以在后续解码中引导目标网格的形变.

### 3.2 基于 FFAdaIN 的网格解码器

DSFFNet 的网格解码器由 3 组 FFAdaIN ResBlock 单元构成, 以目标顶点 $v_{id}$ 作为前向输入, 混合特征 $f_{mix}$ 与目标顶点 $v_{id}$ 作为侧面输入. FFAdaIN 模块的详细架构如图 3(b)所示, 竖直方向表示 FFAdaIN 的前向传播方向. FFAdaIN 的计算过程如下:

$$\begin{aligned}\gamma_{id} &= \text{Conv1d}_{id1}(v_{id}) \\ \delta_{id} &= \text{Conv1d}_{id2}(v_{id}) \\ \gamma_{mix} &= \text{Conv1d}_{mix1}(f_{mix}) \\ \delta_{mix} &= \text{Conv1d}_{mix2}(f_{mix})\end{aligned} \quad (1)$$

$$\begin{aligned}\gamma_{ff} &= \alpha\gamma_{id} + (1-\alpha)\gamma_{mix} \\ \delta_{ff} &= \beta\delta_{id} + (1-\beta)\delta_{mix}\end{aligned} \quad (2)$$

$$FFAdaIN(h_{in}, h_{mix}, v_{id}) = \gamma_{ff}\frac{(h_{in}-\mu_{in})}{\sigma_{in}} + \delta_{ff} \quad (3)$$

为了解决姿态特征在前向传播中的失真问题, DSFFNet 采用了直接学习未失真姿态的方法. 在 FFAdaIN 模块内, 设计了两个侧通道: 混合特征侧通道和目标顶点侧通道, 每个侧通道包括两个一维卷积层. 混合特征侧通道从 $f_{mix}$ 中提取包含姿态信息的混合特征变量 $\gamma_{mix}, \delta_{mix}$, 而目标顶点侧通道从 $v_{id}$ 中提取包含身份信息的身份特征变量 $\gamma_{id}, \delta_{id}$. 两个可学习参数 $\alpha$ 和 $\beta$ 在训练过程中自动调整, 以找到姿态和身份特征融合的最佳比例. 混合特征变量 $\gamma_{mix}, \delta_{mix}$ 与身份特征变量 $\gamma_{id}, \delta_{id}$ 通过 $\alpha$ 和 $\beta$ 加权求和来构建特征融合参数 $\gamma_{ff}$ 和 $\delta_{ff}$, 这些参数用于前向通道的反归一化. FFAdaIN 的前向输入 $h_{in}$ 在经过实例归一化后, 使用 $\gamma_{ff}$ 和 $\delta_{ff}$ 进行反归一化, 从而得到输出. 其中 $\mu_{in}$ 和 $\sigma_{in}$ 是 $h_{in}$ 在每个 Channel 维度上的均值和标准差.

在 NPT 中, 空间自适应实例归一化模块 SPAdaIN 只包含学习目标网格身份特征的侧通道, 而前向通道输入的混合特征虽然包含姿态特征, 但在解码器中经过多次前向传播后姿态特征就会出现失真问题. 此外, 唯一的侧通道也表明 NPT 解码器在前向传播过程中是对目标网格的身份特征进行补偿, 也就是说, NPT 本质上是通过迁移目

标网格的身份特征来间接地实现姿态迁移. 而 FFAdaIN 通过加入混合特征侧通道, 使模块具有了学习源网格姿态特征的能力, 能够在前向传播过程中对失真的姿态特征进行补偿. 由 FFAdaIN 构成的网格解码器以目标顶点作为前向输入, 在前向传播过程中由 FFAdaIN 对目标顶点进行姿态修正, 从而提高姿态迁移的精度. 这种方法有效地解决了姿态迁移中的失真问题, 使得 DSFFNet 在姿态迁移任务上表现更出色.

图 3(c)设计了 FFAdaIN ResBlock. 在 NPT 中 SPAdaIN ResBlock 包含 3 个 SPAdaIN 单元, 取消其中任意一个 SPAdaIN 单元就会导致模型无法收敛到理想范围内. 而本文设计的 FFAdaIN 使得姿态特征在传播过程中避免了失真现象, 因此不需要更多的 FFAdaIN 来构建残差结构以提升网络性能. 事实上, 只使用 2 个 FFAdaIN 就足以使网络性能超过 NPT, 同时保持了较小的网络体积. 这进一步证明了 DSFFNet 在姿态迁移任务上的高效性和性能优势.

### 3.3 损失函数

通过最小化以下损失函数来训练 DSFFNet 以达到最佳性能.

**重建损失** 与之前的许多方法一样[5-7,19], 本文采用重建损失作为模型的优化目标函数. 这一损失函数要求模型生成的输出网格与 Ground Truth 网格具有相同的顶点顺序. 由于输出网格是通过对目标网格进行姿态迁移得到的, 因此目标网格的顶点顺序与输出网格一致. 因此, 需要将 Ground Truth 的顶点按照目标网格的顶点顺序进行排序. 重建损失按照公式(4)计算.

$$L_{rec} = \frac{1}{N}\sum_{i=1}^{N}\left\|x_{pred} - x_{gt}\right\|_2^2 \quad (4)$$

式中 $N$ 为顶点个数, $x_{pred}$ 和 $x_{gt}$ 分别是输出网格和 Ground truth 网格顶点的坐标.

**边长约束损失** 边长约束的作用是避免在姿态迁移过程中出现较大的边长变化, 从而使输出网格更加光滑. 与之前的方法[5-7]不同, 本文认为使用输出网格和 Ground Truth 网格作为损失函数的输入有助于提高姿态迁移的效果. 边长约束损失按照公式(5)计算.

$$L_{edg} = \frac{1}{L}\sum_{i=1}^{L}\left|l_{pred}/l_{gt} - 1\right| \quad (5)$$

式中 $L$ 是边的个数, $l_{pred}$ 和 $l_{gt}$ 分别是输出网格和 Ground Truth 网格的边长.

本文采用公式(6)的联合损失函数对模型进行优化.

$$L = L_{rec} + \lambda L_{edg} \quad (6)$$

式中 $\lambda$ 的值选用 0.0005.

## 4 实 验

### 4.1 实验背景

**数据集** 本文采用了 NPT[5]生成的基于 SMPL[29]的人体网格数据集, 用于 DSFFNet 的训练和评估. 该数据集包含了 16 种身份和 400 种姿态的训练集, 以及 14 种身份和 800 种姿态的验证集. 验证集包括 400 种与训练集相同的已见过的姿态和 400 种未见过的姿态.

本文还使用了 3D-CoreNet[6]生成的基于 SMAL[30]的动物网格数据集. 该动物数据集的训练集包含了 24 种身份和 300 个姿态, 验证集包含了 16 种身份和 600 个姿态, 其中包括 300 种已见过的姿态和 300 种未见过的姿态. 这个数据集用于训练和评估针对动物网格的 DSFFNet 模型, 旨在扩展 DSFFNet 的应用领域, 使其可以处理动物网格数据的姿态迁移任务.

此外, 本文还选用了来自 FAUST[31]和 MultiGarment[32]数据集中的人体网格数据, 用于测试 DSFFNet 的泛化能力以及在不同顶点数的网格上执行姿态迁移任务的性能. 这两个数据集与 SMPL 数据集的姿势和身份不同. FAUST 和 SMPL 的网格都包含 6890 个顶点, 而 MultiGarment 的网格包含 28920 个顶点, 而且穿着服饰. 这些额外的数据集充实了 DSFFNet 的评估实验, 展示了它在多样化网格和场景下的适应能力.

**评估指标** 本文采用了经典的点云评估指标, 包括点状网格欧氏距离(Point-wise Mesh Euclidean Distance, PMD), Chamfer Distance(CD), 以及 Earth Mover's Distance(EMD)[33]来评估模型的性能. 这些指标主要用于衡量模型生成的输出网格与 Ground Truth 网格之间的相似性. 在这些指标中, PMD、CD 和 EMD 的值越小越好, 表示生成的网格与 Ground Truth 的相似度越高, 模型性能越好.

**实施细节** 在数据准备过程中, 源网格和目标网格都是随机选择并打乱了顶点顺序, 这有助于提高模型的泛化能力. Ground Truth 网格是根据源网格的姿态和目标网格的身份来选择的. 在送入模型之前, 所有的网格顶点坐标都要进行归一化.

表 1 定量对比结果

| Dataset | Pose Type | Method | PMD↓(×10⁻⁴) | CD↓(×10⁻⁴) | EMD↓(×10⁻³) |
|---|---|---|---|---|---|
| SMPL [29] | Seen | NPT | 2.39 | 10.61 | 49.02 |
| | | 3D-CoreNet | 1.13 | 3.24 | 23.76 |
| | | **Ours** | **0.57** | **1.37** | **13.40** |
| | Unseen | NPT | 8.60 | 38.72 | 198.18 |
| | | 3D-CoreNet | 4.45 | 14.34 | 96.64 |
| | | **Ours** | **1.78** | **2.18** | **21.44** |
| SMAL [30] | Seen | NPT | 50.67 | 153.97 | 106.89 |
| | | 3D-CoreNet | 18.86 | 48.72 | 83.25 |
| | | **Ours** | **11.32** | **19.52** | **41.73** |
| | Unseen | NPT | 70.28 | 212.39 | 162.48 |
| | | 3D-CoreNet | 25.53 | 75.78 | 132.54 |
| | | **Ours** | **17.96** | **43.29** | **72.46** |

表 2 不同方法的模型大小和平均推理时间

| Method | NPT | 3D-CoreNet | Ours |
|---|---|---|---|
| Size | 23668KB | 95647KB | 36631KB |
| Time | 0.0071s | 0.0152s | 0.0076s |

这个处理过程包括将网格的中心移动到原点并按照边界框的最长边进行坐标缩放，以确保所有坐标值都位于-1 和 1 之间. 模型的训练过程采用了 AdamW 优化器，初始学习率为 1e-3. 每隔 8 个 epoch，学习率按照上一个学习率的 0.8 倍进行衰减. 总共训练了 200 个 epoch. Batch size 设置为 8，训练过程中使用了一块 RTX 3080 12GB 显卡.

### 4.2 定量对比

表 1 总结了 DSFFNet 与现有的 NPT[5]和 3D-CoreNet[6]方法之间的性能对比结果. 在 SMPL 人体网格数据集上，DSFFNet 对于已见过的姿态具有 0.57 的 PMD、1.37 的 CD 和 13.40 的 EMD，对于未见过的姿态具有 1.78 的 PMD、2.18 的 CD 和 21.44 的 EMD. 这些结果在各项指标上都优于 NPT 和 3D-CoreNet. 在 SMAL 动物网格数据集上，DSFFNet 对于已见过的姿态具有 11.32 的 PMD、19.52 的 CD 和 41.73 的 EMD，对于未见过的姿态具有 17.96 的 PMD、43.29 的 CD 和 72.46 的 EMD，同样取得了良好的结果．与排名第二的 3D-CoreNet 相比，DSFFNet 在 SMPL 的未见过姿态方面降低了 2.67、12.16 和 75.20 的评估指标，对于 SMAL 的未见过姿态降低了 7.57、32.49 和 60.08. 虽然 DSFFNet 不能像 3D-CoreNet 一样生成源和目标网格的对应关系，但在姿态迁移的精度上明显优于 3D-CoreNet.

表 2 记录了 NPT、3D-CoreNet 和 DSFFNet 的模型大小和平均推理时间. 模型大小和推理时间是衡量模型性能的重要指标. 可以看出，NPT 模型最小，推理时间最短，而 3D-CoreNet 模型最大，推

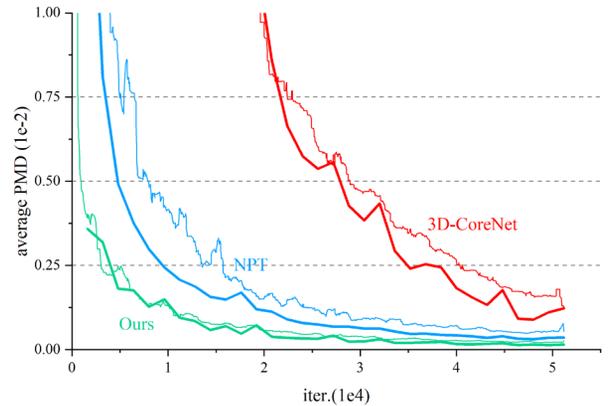

图 4 迭代过程中的误差曲线. 细线为每次迭代的训练误差. 粗线是每个 epoch 的评估误差.

理时间最长. DSFFNet 的模型大小和推理时间介于两者之间，更接近于 NPT 的大小. 与 3D-CoreNet 相比，DSFFNet 的模型大小和推理时间都要小得多，但准确性更高，这表明 DSFFNet 的特征提取效率更高.

图 4 展示了三种模型在 51200 次迭代中的平均 PMD 曲线. 细曲线表示训练过程中每次迭代的误差，粗曲线表示每个 epoch 的验证误差. 可以明显看出，DSFFNet 的收敛速度远快于 NPT 和 3D-CoreNet，而且在接近迭代结束时，训练和验证误差都明显小于它们.

总的来说，DSFFNet 有效提高了姿态迁移的精度和收敛速度. 这是因为模型中的每个 FFAdaIN 模块都能够直接学习到未失真的姿态特征，从而在逐层前向传播过程中对失真的姿态特征进行补偿，进一步提高了姿态迁移的精度，并使模型能够快速收敛. 与此不同，NPT 和 3D-CoreNet 在解码部分没有对失真的姿态特征进行补偿，因而降低了姿态迁移的精度.

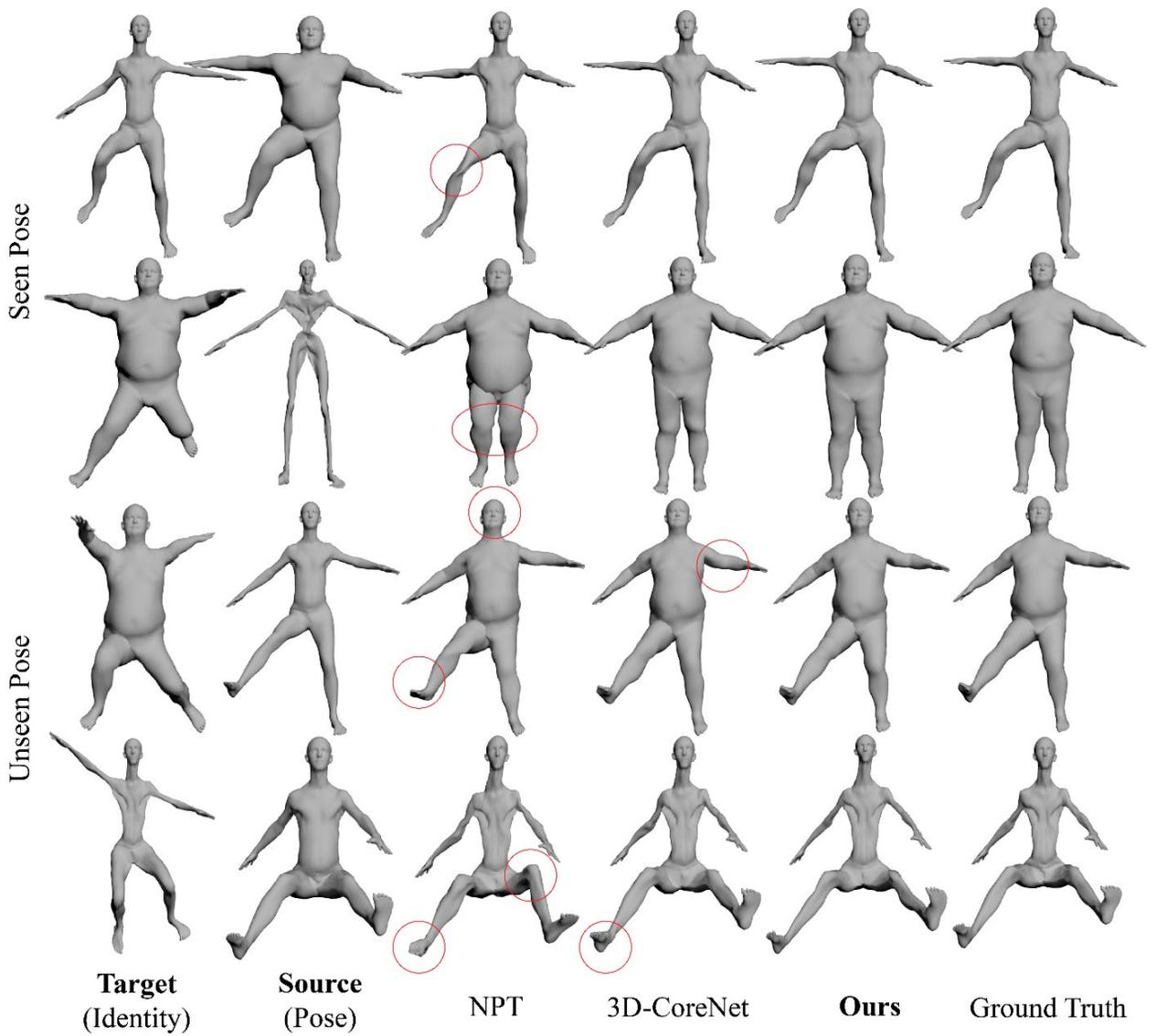

图 5 人体网格的定性比较结果

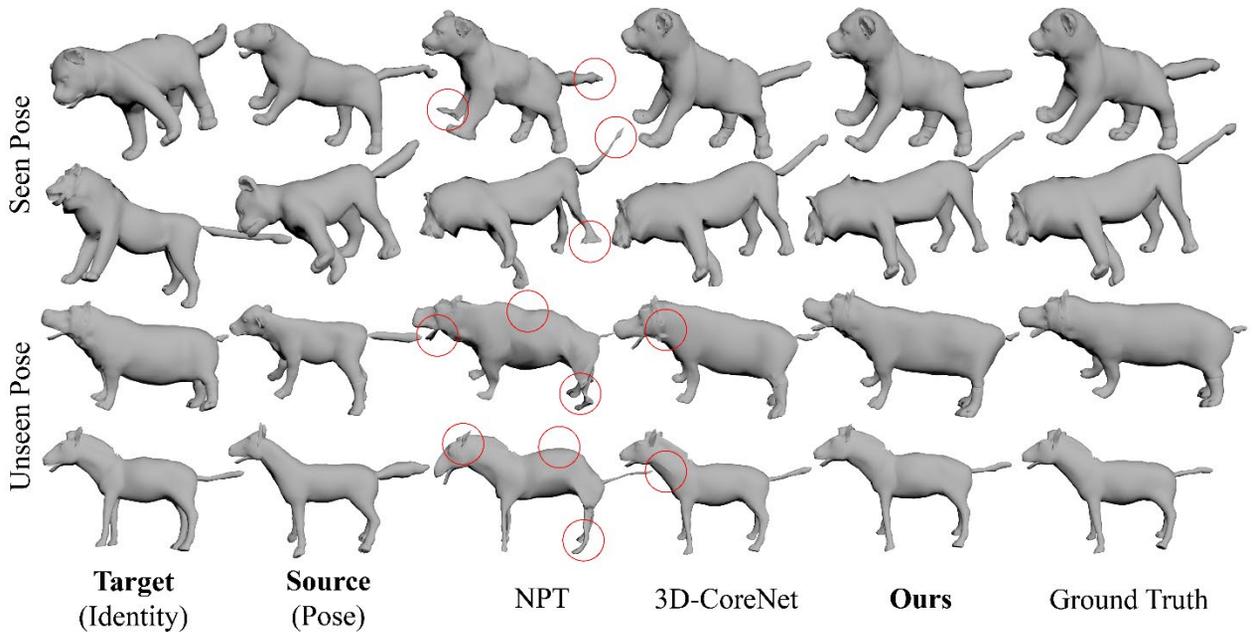

图 6 动物网格的定性比较结果

表 3 消融实验定量对比

| Pose Type | Metrics | w/o FFAdaIN | w/o Side Target | w/o $L_{edg}$ | **Full model** |
|---|---|---|---|---|---|
| Seen | PMD | 7.28 | 2.14 | 0.91 | **0.57** |
|  | CD | 21.23 | 8.33 | 3.67 | **1.32** |
|  | EMD | 119.77 | 34.13 | 22.49 | **13.40** |
| Unseen | PMD | 10.72 | 4.07 | 2.12 | **1.78** |
|  | CD | 49.52 | 18.93 | 7.82 | **2.15** |
|  | EMD | 227.83 | 83.31 | 42.63 | **21.44** |

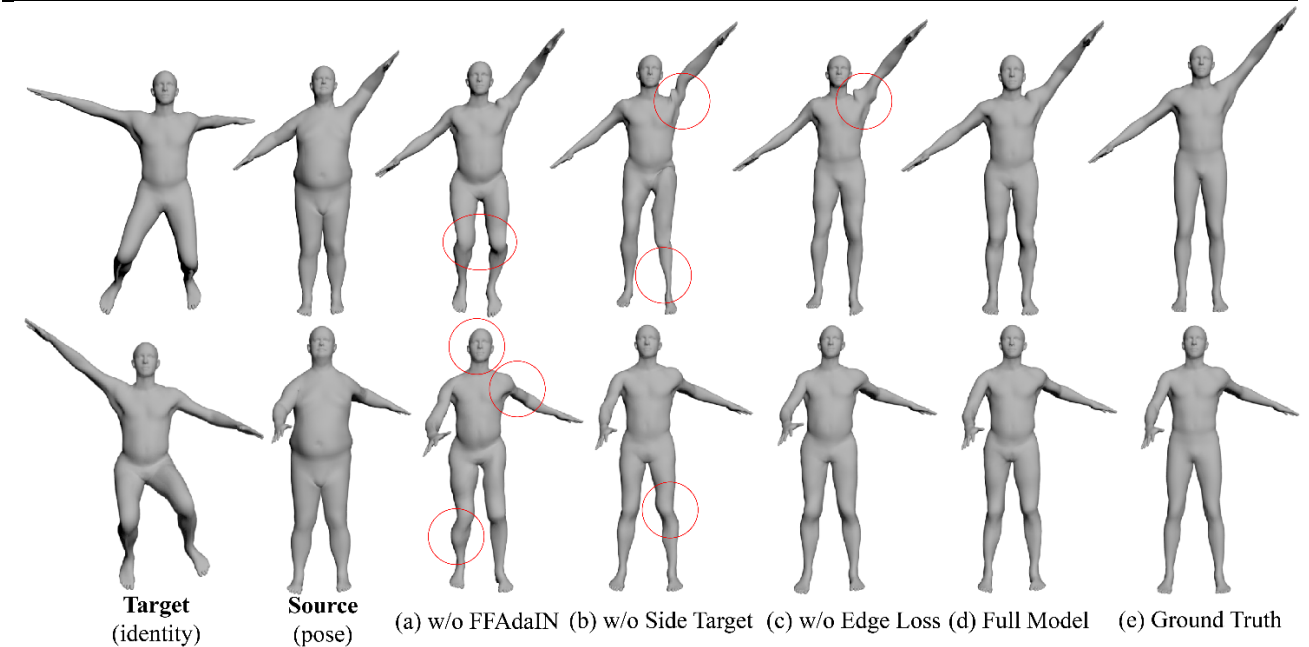

图 7 消融实验定性对比

## 4.3 定性对比

图 5 和图 6 分别展示了在 SMPL 和 SMAL 数据集上的定性对比结果，对比对象包括 NPT[5]、3D-CoreNet[6]和本文提出的 DSFFNet. 在图 5 中，可以看到 NPT 在人体网格上生成了扭曲的头部和扁平的足部，腿部的姿态也出现错误. 3D-CoreNet 的结果与 DSFFNet 类似，但在一些部位存在微弱的扭曲. 在图 6 中，NPT 生成的动物网格存在严重的失真，足部和头部出现扭曲，主体部分也严重偏离源网格的姿态. 3D-CoreNet 的结果在一些部位不够平滑. 相比之下，DSFFNet 很好地保留了网格的局部细节，并将姿态准确地迁移到身份网格上. 这些定性对比结果进一步表明 DSFFNet 在姿态迁移任务上具有明显的优势，能够生成高质量的结果.

## 4.4 消融实验

本文对 DSFFNet 进行了三个不同变体的消融实验，旨在验证模型中各个组成部分的重要性. 表 3 和图 7 展示了这三组实验的定量和定性对比结果. 以下是这些消融实验的方法和结论：

**SPAdaIN 替换 FFAdaIN**：第一个变体FFAdaIN ResBlock 正向传播方向的第一个 FFAdaIN 模块替换为 SPAdaIN 模块. 结果显示，使用 SPAdaIN 替换 FFAdaIN 后，对于已见过的姿势，PMD、CD 和 EMD 的度量值均上升，分别上升了 6.71、19.91 和 106.37. 对于未见过的姿势，这些度量值上升更为显著，分别上升了 8.94、47.37 和 206.39. 这表明本文提出的 FFAdaIN 对提升姿态迁移精度有着至关重要的作用. SPAdaIN 替换 FFAdaIN 后，输出的网格出现多个不光滑的部位，姿态特征在经过 SPAdaIN 时出现失真，导致了精度下降.

**取消目标顶点侧通道**：第二个变体取消了 FFAdaIN 的目标顶点侧通道，以验证是否可以仅依赖混合特征通道来实现较高的精度. 结果显示，取消目标顶点侧通道后，模型仍然能够进行姿态迁移，但相比于完整模型，存在很大的误差. 此时，输出的网格出现了扭曲和不平滑的部位，而腿部变细的现象表明身份特征未能有效保留，说明目标顶点侧通道对保留身份特征具有重要作用.

**取消边长约束损失**：第三个变体取消了边长

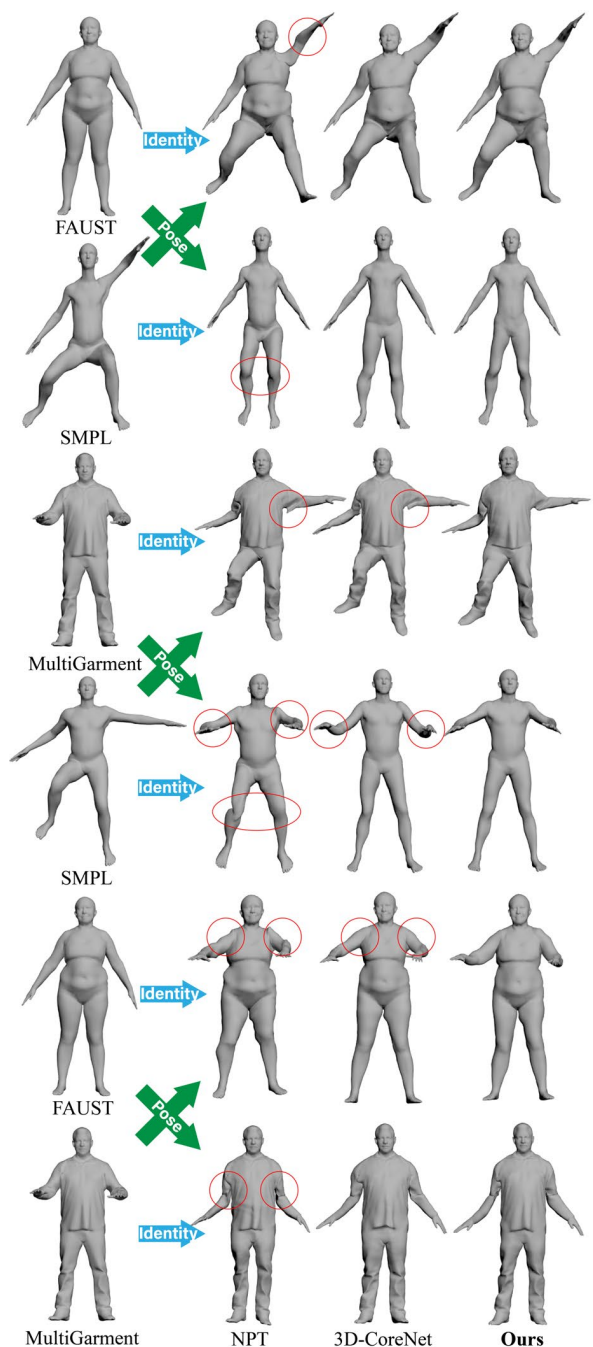

图 8 非 SMPL 复杂网络的姿态迁移结果

约束损失函数, 以验证其对模型精度的重要性. 结果显示, 取消边长约束损失后, 模型的精度下降, 网格出现了部分扭曲和不平滑的现象. 这表明边长约束损失对提高输出网格的平滑度是有意义的.

综合上述结果, 实验验证了 FFAdaIN、目标顶点侧通道和边长约束损失函数在 DSFFNet 中的重要性. FFAdaIN 的作用在于避免姿态特征在传播过程中的失真, 提高姿态迁移精度. 目标顶点侧通道的作用是保留身份特征, 而边长约束损失有助于使输出网格更加平滑. 这些实验结果进一步证明

了 DSFFNet 的有效性和各个组成部分的关键作用.

### 4.5 泛化能力

本文使用 FAUST[31]和 MultiGarment[32]数据集中的人体网格评估 DSFFNet 的泛化能力, 并选择 NPT[5]和 3D-CoreNet[6]进行对比. 如图 8 所示设计了三组实验用于验证 DSFFNet 在处理不同数据集和未见过的身份和姿态时的性能. 以下是这些实验的主要结果和结论:

**FAUST 和 SMPL 的姿态迁移**: 在第一组实验中, NPT 的输出在 FAUST 网格上导致扁平的手臂, 而在 SMPL 网格上未能成功迁移腿部姿态. 这表明 NPT 在处理不同数据集上的泛化性能较差, 容易出现形变和姿态失真. 相比之下, DSFFNet 在这一组实验中表现出色, 成功保留了姿态信息, 表明其对不同数据集的泛化性能更强.

**MultiGarment 和 SMPL 的姿态迁移**: 在第二组实验中, NPT 和 3D-CoreNet 的输出在 MultiGarment 网格上导致了衣服的撕裂, 而在 SMPL 网格上也出现了腿部和手臂的姿态失真. DSFFNet 在这组实验中也表现出色, 成功地进行了姿态迁移, 展现出了强大的泛化能力.

**FAUST 和 MultiGarment 的姿态迁移**: 在第三组实验中, NPT 的输出在 FAUST 网格上产生了扭曲, 而在 MultiGarment 网格上未能正确迁移肩膀处的姿态. DSFFNet 在这一组实验中同样表现出色, 成功地将不同数据集上的姿态迁移到目标网格上, 再次证明了其强大的泛化性能.

总的来说, 这些泛化能力实验结果表明, DSFFNet 相对于 NPT 和 3D-CoreNet 具有更好的泛化性能. DSFFNet 能够有效处理不同数据集、不同身份和不同姿态的网格, 保持高质量的姿态迁移, 这为其在实际应用中的可用性提供了有力支持.

### 4.6 鲁棒性

通过给来自 SMPL 的源网格添加不同幅度的随机噪声, 本文测试了 DSFFNet 在面对噪声数据时的鲁棒性. 图 9 的实验结果显示, 随着噪声幅度的增加, DSFFNet 的输出网格没有受到干扰, 仍然能够准确地完成姿态迁移任务. 这表明 DSFFNet 对于输入数据中的噪声具有很好的鲁棒性, 能够稳定地处理具有噪声的三维模型. 这一特性对于实际应用中处理现实世界中的扫描数据非常重要, 因为这些数据常常存在噪声和不完整性.

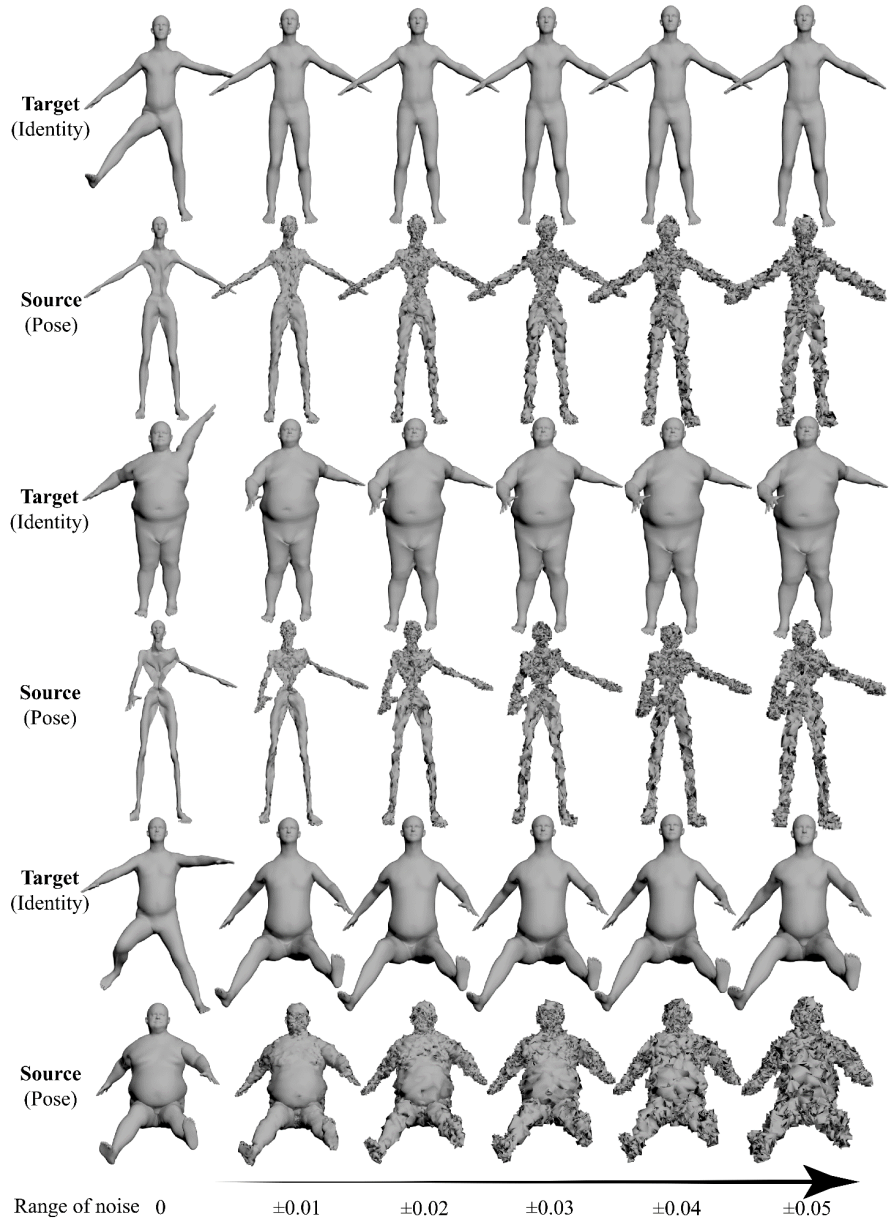

图 9 噪声源网格的姿态迁移结果

综合来看, 鲁棒性实验的结果证明了 DSFFNet 在处理带有噪声的三维模型时的能力, 进一步加强了其在实际场景中的可靠性和实用性.

## 5 结 论

本文提出了一种基于双侧通道特征融合的三维姿态迁移方法 DSFFNet. 为了解决现有方法在姿态迁移过程中的姿态失真问题, 提出一种特征融合自适应实例归一化模块 FFAdaIN, 使网格解码器在前向传播过程中能够对姿态特征进行补偿. 同时, 本文还采用固定长度的一维向量作为姿态编码, 使得 DSFFNet 可以适用于不同顶点数量的网格. 本文在 SMPL、SMAL、FAUST 和 MultiGarment 数据集上进行了大量实验, 结果表明, 本文方法有效解决了姿态失真问题, 提高了姿态迁移精度和训练速度, 相较于现有方法具有更高的效率和更好的实用性.

在未来的工作中, 计划解决 DSFFNet 的几个局限性. 首先, DSFFNet 在处理自接触网格的姿态迁移时存在一定的不足, 需要进一步研究以改进这一方面. 其次, 由于监督训练中 Ground Truth 难以获取, 因此将致力于改进 DSFFNet, 使其具备无监督学习的能力.

为加强防范学术不端行为，**自 2022 年起，投稿时**必须按照投稿模板排版，并在下表中填写所有作者信息．若未按要求填写或所填信息不准确、**未按要求填写工作邮箱、特殊情况未在下表中说明或后续未提供加盖单位公章的说明原件**，将**不予收稿**（此信息仅供编辑部审查使用，发表时不会呈现在论文中）．**注意：工作邮箱是指域名为工作单位的邮箱；例如 QQ, 163, 126, 139, sina, hotmail, gmail 等均非工作单位邮箱，均是公众邮箱（公众邮箱包括但不限于以上列举的邮箱，抱歉无法一一枚举）．**

| 作者姓名<br>（按论文署名顺序填写） | 单位 | 工作邮箱<br>（域名为工作单位的邮箱） | 特殊情况说明<br>（包括单位邮箱涉密不可公开、非员工无单位邮箱等情况请说明） |
|---|---|---|---|
| 刘珏 | 东南大学自动化学院<br>东南大学复杂工程测量与控制教育部重点实验室 | liu_jue@seu.edu.cn<br>电话 17361730126<br>QQ1658959262 | |
| | | | |
| | | | |
| …… | | | |